\documentclass[letterpaper, 10 pt, conference]{ieeeconf}
\IEEEoverridecommandlockouts
\usepackage{float}
\usepackage{cite}
\usepackage{amsmath,amssymb,amsfonts}
\usepackage{graphicx}
\usepackage{booktabs}
\usepackage{multirow}
\usepackage{url}
\usepackage{hyperref}
\usepackage{xcolor}
\usepackage[T1]{fontenc}
\usepackage{fancyvrb}
\usepackage{tcolorbox}
\usepackage{upquote}

\definecolor{jsonkey}{HTML}{0550AE}
\definecolor{jsonstr}{HTML}{0A7A3D}
\definecolor{jsonnum}{HTML}{B35900}
\definecolor{jsonpunct}{HTML}{57606A}
\definecolor{jsonbg}{HTML}{F6F8FA}
\definecolor{jsonframe}{HTML}{D0D7DE}
\newcommand{\jk}[1]{\textcolor{jsonkey}{#1}}
\newcommand{\js}[1]{\textcolor{jsonstr}{#1}}
\newcommand{\jn}[1]{\textcolor{jsonnum}{#1}}
\newcommand{\jp}[1]{\textcolor{jsonpunct}{#1}}

\title{\LARGE \bf
Talk to Act: A Synthetic Multi-Turn Dialogue Benchmark for Human Intent Extraction and Action Planning in Robotics
}

\author{
Deepu Krishnareddy$^{1}$, Mohammad Haziq Khan$^{1}$, Michael Danner$^{2}$, Matthias Rätsch$^{1}$%
\thanks{$^{1}$Reutlingen University, Germany. $^{2}$Bochum University of Applied Sciences, Germany.}%
}
\begin{document}

\maketitle
\thispagestyle{empty}
\pagestyle{empty}

\begin{abstract}
Multi-turn conversational human–robot interaction (HRI) continues to be a challenging problem in robotics, as many existing systems rely on rigid, command-driven interfaces that can be unintuitive for users. Recent advances in Large Language Models (LLMs) provide a promising direction, given their ability to process natural language, generate structured outputs, and support tool-based reasoning, all of which are relevant for interpreting user intent in robotic settings. However, progress in this area is constrained by the limited availability of datasets and benchmarks that capture the nuances of real-world human communication, such as multi-turn dialogue, ambiguity, and context-dependent requests.

In this work, we present a synthetically generated multi-turn dialogue dataset for HRI, annotated with action intents, object references, and task plans across a range of scenarios and user personas. Our primary focus is on evaluating the capability of small, open-weight language models for robotic tasks, particularly in settings where local deployment and efficiency are important. To this end, we frame intent extraction as a tool-calling problem where each predicted action corresponds to a structured function call with arguments grounded in the dialogue and benchmark models ranging from 270M to 8B parameters on their ability to perform intent extraction and action planning without task-specific fine-tuning.

Our results show that these models can achieve reasonable performance on the task even at relatively small scales and without any task-specific fine-tuning. This suggests that compact, locally deployable models already provide a useful baseline for conversational robotic systems. However, the results also highlight clear potential for improvement. We expect that task-specific fine-tuning and more realistic training data will enhance performance further and make such models increasingly suitable for real-world robotic applications.
\end{abstract}

\section{Introduction}

\begin{figure*}[t]
    \centering
    \includegraphics[width=\textwidth]{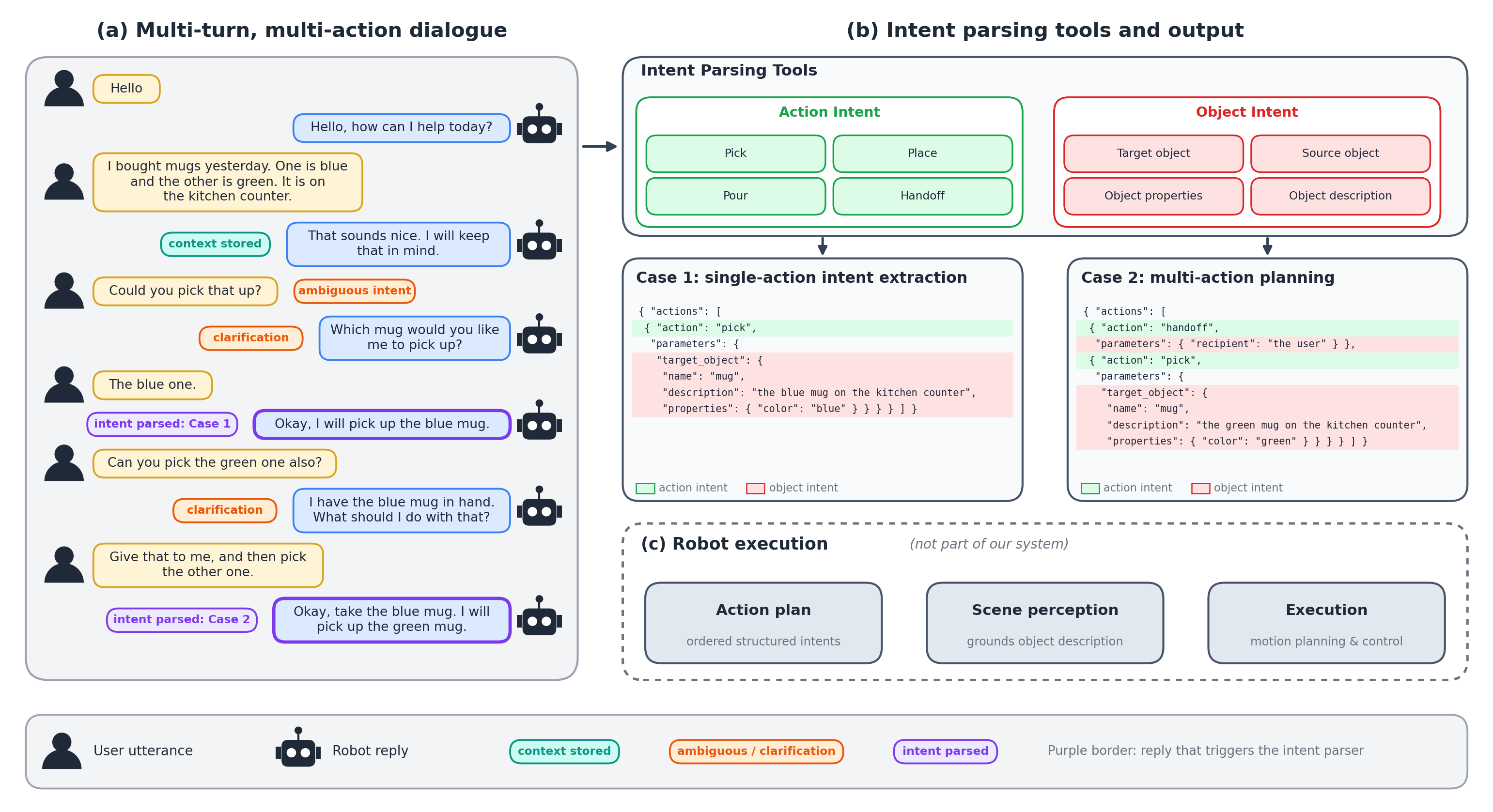}
    \caption{Overview of the proposed HRI framework. A multi-turn conversation between a user and a robot is processed by context-aware intent parsing tools. Ambiguous references (e.g., ``that'', ``the other one'') are resolved using the dialogue context and action history. The parser outputs action intent (pick, place, pour, handoff) and object intent (target object, source object, properties, and description) as structured JSON.}
    \label{fig:overview}
\end{figure*}

Human-Robot Interaction (HRI) has emerged as a critical frontier in robotics, enabling robots to operate alongside humans in shared environments such as assisted living facilities, warehouses, and domestic settings.  As robots move from controlled industrial cells into homes, care environments, and shared workspaces, communication becomes a core capability rather than a peripheral interface. 

In recent years, substantial progress has been made in robotics, particularly in perception and control. Large embodied datasets and generalist robot policies have advanced robotic performance. Vision-language-action models (VLAs) have demonstrated the effectiveness of web-scale pretraining in enhancing semantic generalization in robotic control. However, interaction between humans and robots is still unnatural. A robot that is capable of perceiving objects and executing movements with precision may still struggle when a user makes an underspecified, multistep, or revised request in everyday language.

Nevertheless, natural language is the most intuitive communication medium for people. Previous studies have contrasted natural language with complex graphical interfaces, structured command languages, and formal abstractions, which require training before a user can effectively control a robot. For robots in homes, assistive scenarios, or shared human environments, it is often unrealistic to expect the human to learn a rigid interaction protocol. A more realistic objective is for robots to be able to interpret natural, unconstrained, and potentially messy human language as it is actually spoken. However, unrestricted language is challenging because human requests are often context-dependent, incomplete, and ambiguous.

Recent advancements in large language models (LLMs) have introduced new opportunities for integrating natural language into robotic tasks. Models such as GPT-5, Claude and Gemini demonstrate strong capabilities in common sense reasoning, following instructions, and tool calling — properties that are directly relevant to interpreting user intent and translating it into structured robot actions. LLMs ability to handle compositional instructions and generate structured outputs makes LLMs a promising foundation for the interaction layer. Furthermore, LLMs offer a promising approach to generating synthetic data, enabling the creation of extensive and diverse dialogue corpora that would otherwise be too costly to collect through human annotation.

Although LLMs have great potential in HRI, existing approaches tend to reduce interaction to single-turn intent extraction within narrowly defined task domains  \cite{LAI2025126360}. However, real human communication is multi-turn, context-dependent, and often under-specified. Users revise goals, refer to objects implicitly, and issue instructions that are only interpretable when considering the prior conversational context.

While large frontier models may be attractive for intent understanding due to their strong reasoning capabilities, their deployment in robotic systems is constrained by latency, connectivity requirements, cost, and privacy considerations. In contrast, recent compact open-weight models such as Phi-4, Gemma series, and Qwen series demonstrate that smaller models are becoming increasingly capable at instruction following and structured output generation, making them compelling candidates for on-device or edge-assisted HRI pipelines. Whether such models can reliably handle the full complexity of natural, multi-turn human dialogue in HRI contexts, however, remains an open question.

Alongside language, perception capabilities have also advanced significantly. Modern models such as SAM 3 \cite{carion2025sam3segmentconcepts} and Grounded DINO \cite{liu2023grounding} now support open-vocabulary segmentation and detection driven by natural language queries. Yet current HRI systems rarely leverage these capabilities, leaving a gap between what perception models can accept and what interaction pipelines actually provide. A well-designed dialogue system that extracts structured referential expressions from human conversation could directly bridge this gap, feeding natural language grounded outputs into modern perception models.

A key bottleneck in advancing natural language-based HRI is the lack of datasets that reflect the complexity of real human communication. Existing benchmarks are limited in scale, domain diversity, and conversational depth. To address this, we introduce a synthetically generated multi-turn dialogue dataset for HRI, designed to capture natural, ambiguous, and context-dependent human instructions annotated with action intent, action plans, and object intent. The dataset spans a variety of real-world scenarios and user personas, reflecting the diversity of people and contexts a robot may encounter in practice. We further present an initial benchmark of small open-weight language models on this dataset, exploring their ability to extract user intent and task plans from unconstrained human dialogue in HRI settings.

\section{Related Work}
Prior work on natural-language HRI spans intent recognition methods, conversational embodied datasets, and LLM/VLM-based robot systems.
A recent review of industrial collaborative robotics describes intention recognition as covering rule-based, probabilistic, machine learning, and deep learning approaches utilizing physical, physiological, and contextual cues. The review highlights the importance of understanding human intentions for safe and effective collaboration~ \cite{kekana_review_2025}. Earlier grounded-language work addressed this problem through explicit structured reasoning. Vanzo \textit{et al.} used a combination of discriminative structured learning, distributional semantics, and contextual evidence from the environment to resolve ambiguities in robotic commands\cite{vanzo_grounded_2020}. CoBaIR proposed an explainable Bayesian framework for context-based intention recognition in low-data settings \cite{lubitz_cobair_2023}. In physical HRI, Lai \textit{et al.} inferred user intent from force interactions alone through physical human-robot interaction primitives \cite{lai_user_2022}. These approaches are important because they show that grounding, context, and interpretability are essential, but they are not designed around open-ended, multi-turn conversational language as the primary input for intent extraction and task planning.

Conversation-oriented HRI work pushed the field closer to natural communication, but with different evaluation goals. Papaioannou \textit{et al.} showed that a hybrid chat-and-task dialogue system can improve perceived interaction quality, yet their main emphasis is adaptive dialogue management for engagement rather than structured extraction of intent and task plans from dialogue \cite{papaioannou_hybrid_2017}. In embodied AI, ALFRED established a benchmark for long-horizon household instruction following from language and egocentric vision, but it is fundamentally organized around directive-to-action mapping rather than conversation \cite{shridhar_alfred_2020}. TEACh extends the embodied setting substantially by introducing over 3,000 unconstrained human-human dialogues in simulation, including clarification, correction, and interleaving of language with actions, and defines benchmarks for execution from dialogue history, trajectory from dialogue, and two-agent task completion \cite{padmakumar_teach_2021}. These resources are highly relevant to conversational robotics, but they remain oriented toward embodied execution rather than a standalone benchmark with explicit labels for intent extraction and task-planning supervision.

Recent LLM, VLM and VLA work demonstrates the promise of language models in robotics, but mostly from the perspective of control or full-agent behavior. Gao \textit{et al.} fine-tuned GPT-3.5 to interpret assembly instructions and generate corrective actions in a manufacturing case study \cite{gao_integrating_2024}. RobotGPT addresses the instability of directly using ChatGPT-generated code for manipulation and instead learns a more reliable manipulation agent from ChatGPT supervision \cite{jin_robotgpt_2024}. LLaRA and Hi Robot show that conversation-style supervision and synthetic language data can substantially improve open-ended robot instruction following for vision-language-action policies \cite{li_llara_2025, shi_hi_2025}, while Wang \textit{et al.} develop a tool-augmented industrial HRC agent with memory and reasoning mechanisms for aerospace wire-harness assembly \cite{wang_llm_2026}. Collectively, these papers show that language models can support robot reasoning, planning, and interaction, but they do not directly target the question we study here: whether small language models can be used for human robot communication specifically for extracting human intent and producing structured task plans from natural multi-turn HRI dialogue.

A final line of work is especially important for motivating realistic data design. Xu \textit{et al.} demonstrate that synthetic dialogue generation can scale to 10,000 multi-party conversations for proactive task inference~\cite{xu_simulated_2025}.. However, such datasets focus on diverse dialogue between users and do not address intent extraction or task planning in the context of robotics, leaving a gap that our work directly targets These findings motivate our design choices directly. Our dataset is fully synthetically generated using LLMs and partially verified by human evaluators to ensure quality and naturalness. It consists of multi-turn dialogues spanning diverse scenarios and user personas, with each conversation paired with explicit annotations for action intent, object intent, and task planning, targeting the conversational language layer as a practical problem in HRI.

\section{Method}

\subsection{Synthetic Dataset Generation}
\begin{figure}[t]
    \centering
    \includegraphics[width=\columnwidth]{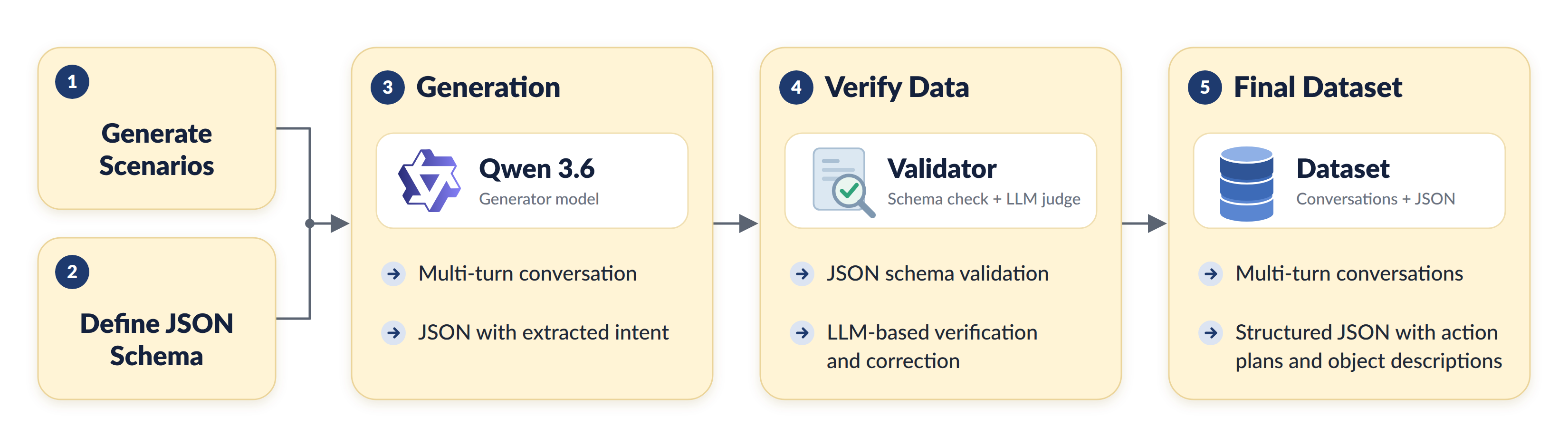}
    \caption{Overview of the synthetic data generation pipeline.}
    \label{fig:data_generation_pipeline}
\end{figure}

We used Qwen 3.6 \cite{qwen36_35b_a3b} as the backbone for synthetic data generation; the full pipeline is illustrated in Figure~\ref{fig:data_generation_pipeline}. 

\

Each conversation consists of at least ten turns between a user and an assistant, spanning everyday chitchat to explicit task-based requests. We formulate intent parsing as a tool-calling problem, where the model must invoke one of four predefined actions, pick, place, pour, and handoff, in response to user requests. Each conversation contains more than one task request, requiring the model to plan multiple actions per conversation while maintaining state across turns. For example, if the robot has already picked an object, it cannot pick again until it has placed it, and the model must track this state throughout the dialogue. The dataset covers a range of scenarios from household tasks to warehouse settings, and includes diverse user personas to reflect realistic interaction conditions. Every generated sample is validated against a predefined JSON schema and then verified and corrected by an LLM to ensure consistent, well-formed annotations. In total, we collected 977 conversations. We release the dataset publicly to facilitate future research in this area.\footnote{Dataset available at: \url{https://huggingface.co/datasets/deepukr007/hri_human_intent_from_conversation}}

\subsection{Action Schema and Object Intent}
We define the following four actions: pick, place, pour, and handoff. Each action is represented by a structured JSON object containing parameters that describe the action type and the objects involved. Object parameters include a name and a full grounding phrase extracted from the conversation. Optional properties include color, shape, material, and texture. The place action also captures the spatial relationship between the target and a reference object. This structured representation is directly motivated by downstream perception requirements. The object name can serve as a query for closed-vocabulary detectors, such as YOLO. Meanwhile, the grounding description which is a rich, natural language phrase derived from the conversation that can be used as a prompt for open-vocabulary perception models, such as Grounded DINO \cite{liu2023grounding} and SAM 3 \cite{carion2025sam3segmentconcepts}, which accept free-form text input. Thus, the extracted object intent acts as a bridge between natural human dialogue and modern perception pipelines without requiring manual reformulation of user instructions. An overview of the full pipeline is illustrated in Fig.~\ref{fig:overview}.

\subsection{Data Verification and Correction}
The generated conversations were reviewed for logical inconsistencies and minor errors using an LLM-based evaluator based on Qwen 3.6  and corrector pipeline, where a correction prompt was used to identify and fix state tracking violations and incoherent action sequences. After this automated correction pass, the data quality was found to be satisfactory for benchmarking purposes.

\subsection{Test Case Construction and Evaluation}
The dataset is split into training and test sets using a random 80/20 split at the conversation level. The training set is retained for future fine-tuning of small language models. Test cases are generated from the test set conversations, where each test case represents the conversation up to the next tool call. Subsequent test cases from the same conversation include the history of prior turns and ground truth responses, allowing evaluation of model performance under increasing context lengths. We evaluate a range of open-weight small language models spanning from 270 million to 8 billion parameters, assessing their ability to extract user intent and generate correct action plans without any task-specific fine-tuning.

\section{Results}

\subsection{Dataset Statistics}

 The dataset comprises 977 conversations generated across 40 unique combinations of user persona, tone, and environment. Environments include homes, old age homes, restaurants, offices, supermarkets, educational institutions, airports, and warehouses, covering a broad range of realistic HRI scenarios. Figure \ref{fig:json-example} shows an example conversation from the dataset, illustrating how ambiguity is resolved across turns before the model issues a structured tool call with full object grounding information.

 \begin{figure}[]
\begin{SaveVerbatim}[commandchars=\\\{\}, fontsize=\scriptsize]{jsonexample}
\jp{\{}
  \jk{"conversation_id"}: \js{"hri_home_office_0058"},
  \jk{"scenario"}: \js{"Remote professional maintains home office}
    \js{humidity and organizes documents, resolving object}
    \js{ambiguity for water jug and notebook, and managing state}
    \js{for placing a report."},

  \jk{"chat_history"}: \jp{[}
    \jp{\{}\jk{"role"}: \js{"user"},
     \jk{"content"}: \js{"Hey. Are you online and ready for the sync}
       \js{in five minutes?"}\jp{\}},

    \jp{\{}\jk{"role"}: \js{"assistant"},
     \jk{"content"}: \js{"I am online and ready. How are you doing}
       \js{today?"}\jp{\}},

    \jp{\{}\jk{"role"}: \js{"user"},
     \jk{"content"}: \js{"Tired. I was up late finishing the Q3}
       \js{report. The air in here feels dry too."}\jp{\}},

    \jp{\{}\jk{"role"}: \js{"assistant"},
     \jk{"content"}: \js{"I understand you need to address the}
       \js{humidity. What would you like me to do regarding the}
       \js{humidifier?"}\jp{\}},

    \jp{\{}\jk{"role"}: \js{"user"},
     \jk{"content"}: \js{"It's empty. Pick the water jug on the side}
       \js{table and fill it up."}\jp{\}},

    \jp{\{}\jk{"role"}: \js{"assistant"},
     \jk{"content"}: \js{"Could you clarify which water jug you are}
       \js{referring to?"}\jp{\}},

    \jp{\{}\jk{"role"}: \js{"user"},
     \jk{"content"}: \js{"The large blue one. The small clear one is}
       \js{for drinking water only."}\jp{\}},

    \jp{\{}\jk{"role"}: \js{"assistant"}, \jk{"content"}: \js{""},
     \jk{"tool_calls"}: \jp{[}\jp{\{}\jk{"type"}: \js{"function"}, \jk{"function"}: \jp{\{}
       \jk{"index"}: \jn{1}, \jk{"name"}: \js{"pour"}, \jk{"arguments"}: \jp{\{}
         \jk{"source_object_name"}: \js{"water jug"},
         \jk{"source_object_description"}: \js{"large blue water jug}
           \js{on side table"},
         \jk{"source_object_properties_color"}: \js{"blue"},
         \jk{"target_object_name"}: \js{"humidifier"},
         \jk{"target_object_description"}: \js{"humidifier on side}
           \js{table"}\jp{\}}\jp{\}}\jp{\}}\jp{]}\jp{\}},

    \jp{...}
  \jp{]}
\jp{\}}
\end{SaveVerbatim}
\begin{tcolorbox}[colback=jsonbg, colframe=jsonframe, boxrule=0.4pt, arc=3pt,
                  boxsep=0pt, left=3pt, right=3pt, top=3pt, bottom=3pt]
\resizebox{\linewidth}{!}{\BUseVerbatim{jsonexample}}
\end{tcolorbox}
\caption{Example conversation \texttt{hri\_home\_office\_0058}.}
\label{fig:json-example}
\end{figure}


\subsection{Evaluation Metrics}

We evaluate model performance using two groups of metrics. 

The first is \textit{Goal Achievement Accuracy}, which measures whether a complete task was accomplished with the correct action plan, in the correct order, and without introducing extra or missing steps. Formally, for a conversation with a ground truth action sequence $A^* = (a_1^*, a_2^*, \ldots, a_n^*)$ and a predicted sequence $\hat{A} = (\hat{a}_1, \hat{a}_2, \ldots, \hat{a}_m)$, goal achievement accuracy is defined as:

\begin{equation}
    \text{Goal Achievement Accuracy} = \frac{1}{N} \sum_{i=1}^{N} \mathbf{1}[\hat{A}_i = A_i^*]
\end{equation}

where $N$ is the total number of instances of task request.

The second group consists of \textit{Action Extraction Metrics} computed at the individual action level across all conversations. For each predicted action, we compute precision, recall, and accuracy where a true positive is a correctly predicted action, a false positive is an action predicted but absent from the ground truth, and a false negative is a ground truth action that was not predicted.




%

\subsection{Benchmark Results}

We evaluate five small open-weight language models out of the box without any task-specific fine-tuning: FunctionGemma (270M), Gemma4:E2B, Gemma4:E4B, Ministral3:3B, and Ministral3:8B, where E denotes effective parameters. Results are reported in Table~\ref{tab:results}.

\begin{table}[t]
\centering
\caption{Benchmark Results of Small Language Models on the HRI Intent Extraction Dataset}
\label{tab:results}
\renewcommand{\arraystretch}{1.2}
\setlength{\tabcolsep}{3pt}
\fontsize{7}{8}\selectfont
\begin{tabular}{lccccc}
\toprule
\multirow{2}{*}{\textbf{Model}} & \textbf{Goal} & \multicolumn{4}{c}{\textbf{Action Level}} \\
\cmidrule(lr){2-2} \cmidrule(lr){3-6}
 & \textbf{Acc. (\%)} & \textbf{Prec. (\%)} & \textbf{Rec. (\%)} & \textbf{F1 (\%)} & \textbf{Acc. (\%)} \\
\midrule
FunctionGemma-270M & 32.3 & 54.7 & 51.7 & 53.2 & 36.2 \\
Gemma4:E2B         & 55.9 & 74.1 & 58.8 & 65.6 & 48.8 \\
Gemma4:E4B         & \textbf{63.6} & \textbf{74.5} & \textbf{74.0} & \textbf{74.2} & \textbf{59.0} \\
Ministral3:3B      & 41.1 & 53.9 & 51.4 & 52.6 & 35.7 \\
Ministral3:8B      & 49.1 & 63.3 & 67.8 & 65.5 & 48.7 \\
\bottomrule
\end{tabular}
\end{table}

\begin{figure}[htbp]
    \centering
    \includegraphics[width=\linewidth]{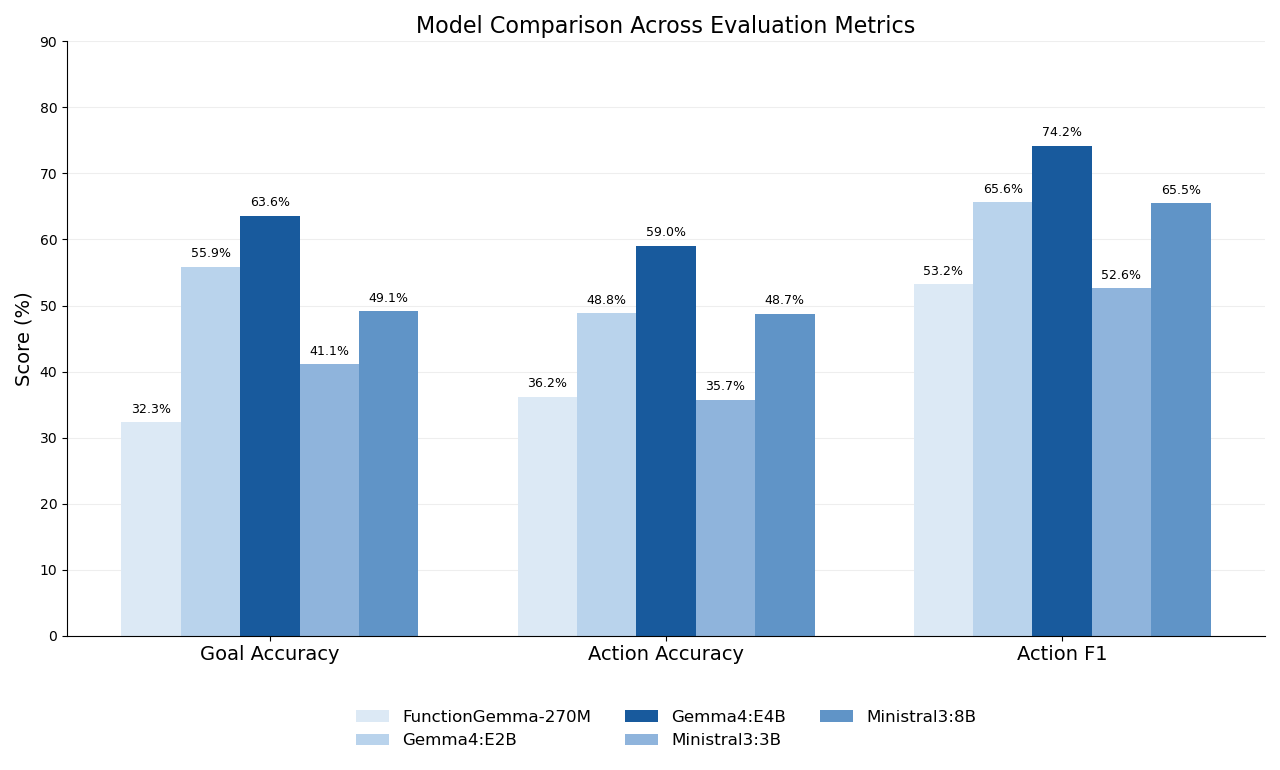}
    \caption{Comparison of five models (FunctionGemma-270M, Gemma4:E2B, Gemma4:E4B, Ministral3:3B, and Ministral3:8B) on Goal Accuracy, Action Accuracy, and Action F1. Gemma4:E4B scores highest on all three metrics (63.6\%, 59.0\%, and 74.2\%). FunctionGemma-270M scores lowest on Goal Accuracy and Action Accuracy, and Ministral3:3B scores lowest on Action F1.}
    \label{fig:model-comparison}
\end{figure}

All models were evaluated using Ollama as the local inference backend, ensuring consistent and reproducible deployment conditions across models. Gemma4:E4B achieved the best performance across all metrics, with a goal achievement accuracy of 63.6\% and an action-level F1 of 74.2\%, demonstrating that a 4 billion effective parameter model can handle a meaningful portion of natural multi-turn HRI dialogue without any task-specific fine-tuning. Notably, FunctionGemma at only 270 million parameters produced competitive results relative to its size, suggesting that models explicitly trained for function calling retain useful structure even in complex conversational settings. Overall, the results are encouraging given that no fine-tuning was performed on any of the evaluated models as shown in Figure\ref{fig:model-comparison} . The scores reflect the inherent difficulty of the task, where conversations are multi-turn, ambiguous, and require state tracking across actions, and suggest that targeted fine-tuning on this dataset has strong potential to further improve performance, which we leave for future work.

\section{Conclusion}
We presented a synthetically generated, multi-turn dialogue dataset for human-robot interaction (HRI), designed to reflect the complexity and naturalness of real human communication. Each conversation is annotated with structured action intent, object intent and task plans, providing a benchmark for evaluating language models on intent extraction and task planning in robotic settings. We evaluated five small, open-weight language models without any task-specific fine-tuning, demonstrating that compact, edge-deployable models can achieve promising performance on this challenging task. These results suggest that small language models are a viable option for on-device HRI pipelines and that fine-tuning these models on this dataset could significantly improve performance.

There are several areas for future work. Firstly, the dataset will be expanded to include more conversations, additional scenarios and a broader range of user personas. Secondly, we intend to fine-tune small language models on the training set in order to evaluate the potential benefits of task-specific adaptation. Thirdly, a wider range of models will be evaluated to provide a more comprehensive benchmark. Fourthly, we will incorporate similarity-based metrics for evaluating object description quality to provide a more detailed assessment of object intent extraction. Fifthly, we intend to incorporate robot state feedback into the dialogue pipeline to enable the robot to communicate its current state naturally, allowing for more grounded and coherent multi-turn interactions. Finally, we plan to incorporate vision and other modalities alongside conversational intent extraction to move towards a fully multimodal human-robot interaction (HRI) system.

\bibliographystyle{IEEEtran}
\bibliography{references}

\end{document}